\documentclass[pdflatex,sn-basic,iicol]{sn-jnl}

\jyear{2022}%

\theoremstyle{thmstyleone}%
\theoremstyle{thmstyletwo}%

\theoremstyle{thmstylethree}%

\algrenewcommand\algorithmicrequire{\textbf{Input:}}
\algrenewcommand\algorithmicensure{\textbf{Output:}}
\newcommand{\removelatexerror}{\let\@latex@error\@gobble}
\definecolor{tabhighlight}{HTML}{e5e5e5}
\definecolor{dkgreen}{rgb}{0,0.6,0}
\definecolor{gray}{rgb}{0.5,0.5,0.5}
\definecolor{mauve}{rgb}{0.58,0,0.82}

\begin{document}

\title[CAR-FT]{Context-Aware Robust Fine-Tuning}


\author*[1]{\fnm{Xiaofeng} \sur{Mao}}\email{mxf164419@alibaba-inc.com}

\author[1]{\fnm{Yufeng} \sur{Chen}}\email{yuefeng.chenyf@alibaba-inc.com}

\author[2]{\fnm{Xiaojun} \sur{Jia}}\email{jiaxiaojun@iie.ac.cn}

\author[1]{\fnm{Rong} \sur{Zhang}}\email{stone.zhangr@alibaba-inc.com}

\author[1]{\fnm{Hui} \sur{Xue}}\email{hui.xueh@alibaba-inc.com}

\author[3]{\fnm{Zhao} \sur{Li}}\email{lzjoey@gmail.com}

\affil*[1]{Alibaba Group, \orgaddress{\city{Hangzhou}, \postcode{310023}, \state{Zhejiang}, \country{China}}}

\affil[2]{Institute of Information Engineering, Chinese Academy of Sciences, \orgaddress{\city{Beijing}, \country{China}}}

\affil[3]{Zhejiang University, \orgaddress{\city{Hangzhou}, \postcode{310027}, \state{Zhejiang}, \country{China}}}


\abstract{Contrastive Language-Image Pre-trained (CLIP) models have zero-shot ability of classifying an image belonging to ``$\mathtt{[CLASS]}$'' by using similarity between the image and the prompt sentence “a $\mathtt{[CONTEXT]}$ of $\mathtt{[CLASS]}$”. Based on exhaustive text cues in ``$\mathtt{[CONTEXT]}$'', CLIP model is aware of different contexts, e.g. background, style, viewpoint, and exhibits unprecedented robustness against a wide range of distribution shifts. However, recent works find further fine-tuning of CLIP models improves accuracy but sacrifices the robustness on downstream tasks. We conduct an empirical investigation to show fine-tuning will corrupt the context-aware ability of pre-trained CLIP features. To solve this problem, we propose \emph{Context-Aware Robust Fine-tuning (CAR-FT)}. CAR-FT regularizes the model during fine-tuning to capture the context information. Specifically, we use zero-shot prompt weights to get the context distribution contained in the image. By minimizing the Kullback-Leibler Divergence (KLD) between context distributions induced by original/fine-tuned CLIP models, CAR-FT makes the context-aware ability of CLIP inherited into downstream tasks, and achieves both higher In-Distribution (ID) and Out-Of-Distribution (OOD) accuracy. The experimental results show CAR-FT achieves superior robustness on five OOD test datasets of ImageNet, and meanwhile brings accuracy gains on nine downstream tasks. Additionally, CAR-FT surpasses previous Domain Generalization (DG) methods and gets 78.5\% averaged accuracy on DomainBed benchmark, building the new state-of-the-art.}

\keywords{Pre-trained Models, CLIP, Fine-tuning, Robustness}



\maketitle

\begin{figure*}[t]
    \centering
    \includegraphics[width=1.0\textwidth]{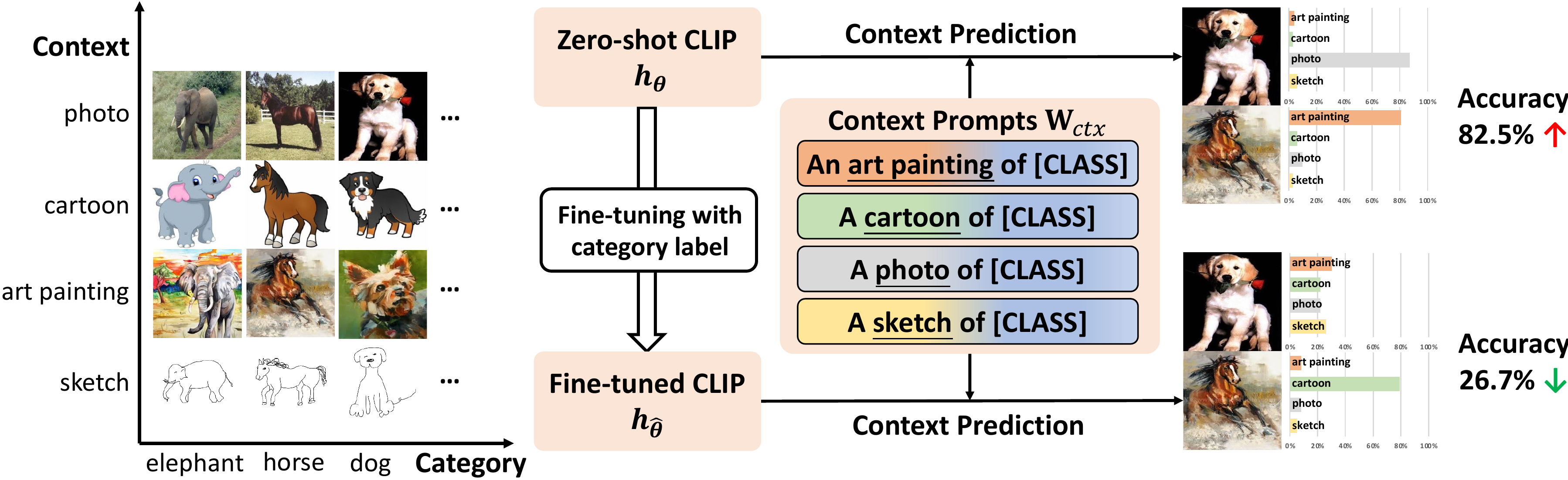}
    \caption{We use PACS dataset to create a task of recognizing context from input images. After fine-tuning CLIP with category label, the context recognition accuracy dropped from 82.5\% to 26.7\%. }
    \label{fig:study}
\end{figure*}

\section{Introduction} \label{sec:intro}
With rough category-level annotation, traditional visual models~\citep{he2016deep,krizhevsky2017imagenet} only focus on the task-related object for classification, and neglect the rest part, i.e. the context or domain information of the image. This paradigm has a natural disadvantage, where the model behaviour becomes unstable and unpredictable when facing abnormal inputs with context or domain out of training set~\citep{he2021towards,zhang2022nico++,moreno2012unifying}. Even high In-Distribution (ID) accuracy can be reached, the giant degeneration of performance still happens on Out-Of-Distribution (OOD) data~\citep{taori2020measuring,hendrycks2021many}. 
Recently, Contrastive Language-Image Pre-training (CLIP)~\citep{radford2021learning} opens up a bright future for narrowing the gap between ID and OOD performance. Instead of category label, CLIP adopts exhaustive text description as supervision to learn visual features. Guided by precise description, CLIP can capture visual concepts not only for classified object, but also other content in the image, e.g. background, style, viewpoint, etc. We collectively refer them as \textit{context}. Such a context-aware visual feature helps the generalization to any other domains or tasks. 
Although CLIP has shown superior zero-shot performance, a supervised fine-tuning is still necessary for yielding further improvement on a specific downstream task. However, several works~\citep{kumar2021fine,wortsman2022robust} have pointed fine-tuning can distort pre-trained features, and make CLIP lose its power on robustness and generalization.~\citet{andreassen2021evolution} explored several fine-tuning approaches but found that they are hard to improve robustness at high accuracy. How to robustly fine-tune pre-trained models is increasingly important and still an open problem till now. Although existing methods add pre-step of linear probing~\citep{kumar2021fine} or post-step of weight averaging~\citep{wortsman2022robust} to prevent the decline caused by fine-tuning, their complex procedure introduces additional computation or heuristic hyper-parameters which is adapted to specific task and has limited versatility.

In this paper, we explore a simple yet effective fine-tuning method from the perspective of context. We first conduct an empirical investigation to show that fine-tuning leads pre-trained models to forget old knowledge about context catastrophically. Detailly, we construct a task of prompt-based~\citep{petroni2019language,liu2021pre} zero-shot context recognition in Figure~\ref{fig:study}. Original CLIP model has initial ability of recognizing a type of context with 82.5\% accuracy. However, such ability is suddenly vanishing with a few epochs of fine-tuning, which appears as a dramatic drop of context recognition accuracy. As a consequence, the fine-tuned model cannot take advantage of context-awareness for classifying images with different contexts, leading to underperformance on OOD classification. 

\begin{figure*}[t]
    \centering
    \includegraphics[width=1.0\textwidth]{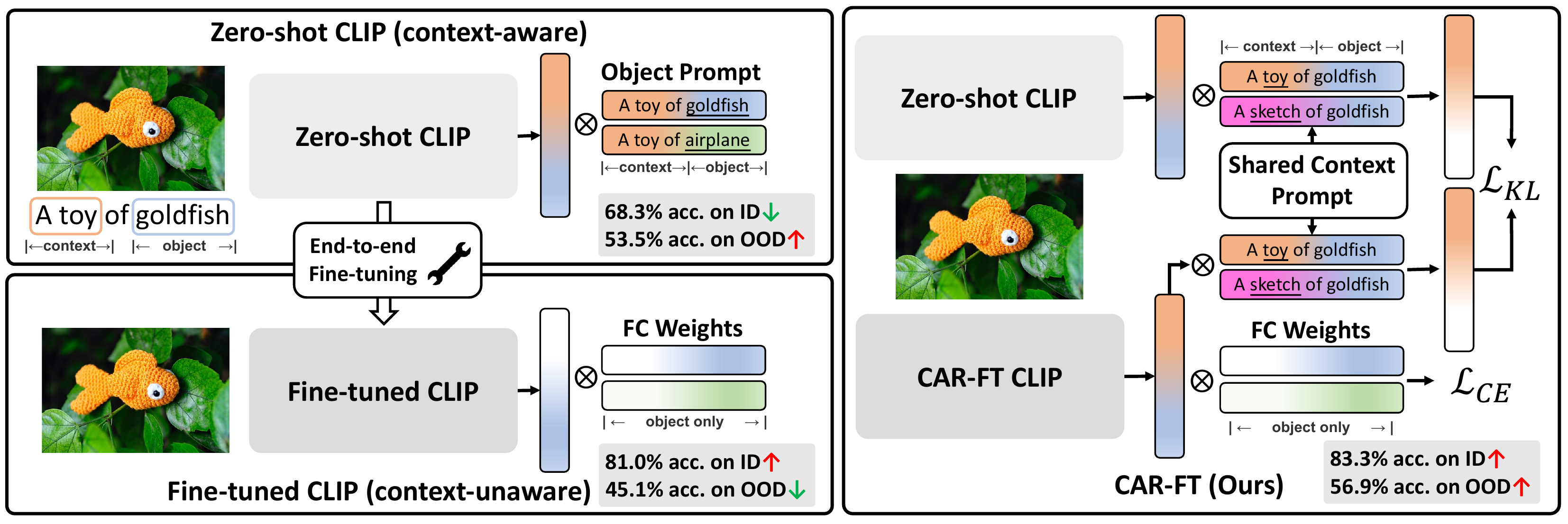}
    \caption{\textbf{(Top left)} Prompt-based zero-shot classification by CLIP. \textbf{(Bottom left)} Fine-tuned CLIP with linear classifier on downstream tasks. \textbf{(Right)} The procedure of our CAR-FT. }
    \label{fig:method}
\end{figure*}

To alleviate such effect, we propose Context-Aware Robust Fine-tuning (CAR-FT), a novel fine-tuning paradigm for pre-trained models. CAR-FT regularizes the model during fine-tuning to capture the context information. However, since common datasets do not always provide annotation of image context, CAR-FT borrows the zero-shot ability of CLIP models to extract the context distributions in the input images. Specially, context distributions is a predicted probability vector of context prompt, which has the same form of ``a $\mathtt{CONTEXT]}$ of a $\mathtt{[CLASS]}$'' but is used for classifying ``$\mathtt{[CONTEXT]}$'' of the image instead of ``$\mathtt{[CLASS]}$''. In this way, we can restrict the fine-tuned feature to encode useful context information by minimizing the Kullback-Leibler Divergence (KLD) between context distributions induced by zero-shot/fine-tuned CLIP models. The detailed procedure is shown in Figure~\ref{fig:method}. 

Benefitting from the context-aware feature, CAR-FT can improve both ID and OOD accuracy on downstream tasks. Thorough experiments are conducted to validate the effectiveness of our method. Concretely, CAR-FT achieves 83.3\% ID accuracy and 56.9\% averaged OOD accuracy on ImageNet classification~\citep{deng2009imagenet}, which is 2.1\% and 8.2\% higher than fine-tuning baseline respectively. Other than ImageNet, CAR-FT brings accuracy gains on nine downstream tasks. As an application to Domain Generalization (DG), CAR-FT gets 78.5\% averaged accuracy on DomainBed~\citep{gulrajani2020search} benchmarks, surpassing existing DG methods and building the new state-of-the-art.  

Our main contributions are in three aspects:
\begin{enumerate}
	\item We empirically demonstrate that the initial context-aware ability of pre-trained models will be corrupted by downstream fine-tuning. As a consequence, models after fine-tuning cannot encode useful image contexts and underperform on OOD classification.
 
	\item We propose a novel fine-tuning paradigm of pre-trained models called Context-Aware Robust Fine-tuning (CAR-FT). It regularizes the model during fine-tuning also encoding context information, and makes the context-aware ability inherited into downstream tasks.
 
	\item The experimental results show CAR-FT achieves superior robustness on five OOD test datasets of ImageNet, and meanwhile brings accuracy gains on nine downstream tasks. Additionally, CAR-FT surpasses previous Domain Generalization (DG) methods and gets 78.5\% averaged accuracy on DomainBed benchmark, building the new state-of-the-art.
\end{enumerate}

\section{Related Work} \label{sec:related}

\textbf{Contrastive Language-Image Pre-training Models.} CLIP~\citep{radford2021learning} has demonstrated that dual-encoder models pre-trained with contrastive objectives on massive image-language pairs can learn generic visual representations. Such representations have zero-shot transfer ability to a variety of downstream classification tasks via prompting. Moreover, it also exhibits remarkable robustness under multiple natural distribution shifts. After the initial success, subsequent works propose further improvement on CLIP framework containing scaling up tasks~\citep{pham2021combined,jia2021scaling}, using pre-trained visual encoders~\citep{zhai2022lit}, combining sub-task of image captioning~\citep{yu2022coca} or expanding more data format~\citep{yuan2021florence}. As the performance of CLIP has a strong correlation with its used datasets~\citep{fang2022data}, there are some efforts~\citep{schuhmannlaion,thomee2016yfcc100m} to create plentiful and useful image-text pairs and make them open to the community. Various opened pre-trained models on these datasets have been used in our paper. Our CAR-FT are applicable to most CLIP based foundation models which support prompting. We also compare the effectiveness of CAR-FT on different backbone scales, datasets and other CLIP variants.  

\noindent\textbf{Robustness under Distribution Shifts.} Practical machine learning systems require the stability under distribution shifts. Previous methods improve the OOD robustness by exploring richer data augmentation~\citep{hendrycks2019augmix,hendrycks2021many}, adversarial training mechanism~\citep{xie2020adversarial,mao2022enhance,herrmann2022pyramid}, advanced network design~\citep{paul2022vision,mao2022towards,bai2021ood,wang2022can} or regularization for a flatter minimum~\citep{cha2021swad,foret2020sharpness}. However, although many advanced techniques are proposed, the giant gap between ID and OOD accuracy still exists~\citep{miller2021accuracy}, which means the robustness of deep models is still far from satisfactory. Fortunately, CLIP~\citep{radford2021learning} contributes a big step to close this gap. It opens a new feasible solution for general robustness by contrastive pre-training on web-scale language-image pairs. Though scaling the model size and training data size of CLIP, powerful foundation models~\citep{pham2021combined,jia2021scaling} can be bulit for achieving state-of-the-art on most OOD benchmarks. This work relies on contrastive language-image pre-trained models, which can provide sufficient context prior of image. Such context knowledge is used by our CAR-FT as possible for improving the generalization on downstream tasks.

\noindent\textbf{Robust Fine-tuning of Pre-trained Models.} Fine-tuning a pre-trained model has become the de facto standard for doing transfer learning in the field of Computer Vision (CV) or Natural Language Processing (NLP). With explosive growth of the capability of large-scale pre-trained models~\citep{radford2021learning}, there are increasing attention to the advanced techniques of fine-tuning. Methods improving the fine-tuning usually aims at higher accuracy on downstream tasks~\citep{ge2017borrowing,guo2019spottune}. However, the robustness and generalization of downstream models are not extensively studied. Some works~\citep{kumar2021fine,wortsman2022robust,andreassen2021evolution} have investigated the evolution of OOD robustness during downstream fine-tuning. They find the effective robustness of pre-trained models will gradually vanish during fine-tuning. LP-FT~\citep{kumar2021fine} proposes to solve this problem by a two-step strategy of first linear-probing to find a good classification head and then full fine-tuning for reducing the distortion of robust pre-trained features. WiSE-FT~\citep{wortsman2022robust} adopts weight-space ensemble of original pre-trained weights and fine-tuned weights to achieve both higher ID and OOD accuracy. This work also aims at robustness and accuracy on downstream task, but differently, we modify the fine-tuning methods by simply adding a loss term, without introducing extra training cost or heuristic hyper-parameters for specific tasks.

\begin{algorithm*}[ht]
\caption{Pseudo code of CAR-FT}
\begin{algorithmic}[1]
\Require 
Pre-trained image encoder $h_{\theta}$ and text encoder $g_{\phi}$; Text prompts $\mathcal{T}$.
\Ensure
Fine-tuned image encoder weights $\hat{\theta}$ and classification weights $\textbf{W}_{f}$.
\State Compute $\textbf{W}_{cls}$ based on $\mathcal{T}$ using Equation~(\ref{eq:obj_prompt})
\State Compute $\textbf{W}_{ctx}$ based on $\mathcal{T}$ using Equation~(\ref{eq:ctx_prompt})
\State Fix parameters $\theta$, $\phi$ and $\textbf{W}_{ctx}$
\State $\hat{\theta} \leftarrow \theta $, $\textbf{W}_{f} \leftarrow \textbf{W}_{cls}$

\For{\texttt{each training steps}}
\State Sample a mini-batch images $x$ with
labels $y$
\State Get reference context distribution of zero-shot model: $p_{ctx}(x;\theta) \leftarrow \text{Softmax}(\textbf{W}_{ctx}^\top h_{\theta}(x))$
\State Get predicted context distribution of fine-tuned model: $p_{ctx}(x;\hat{\theta}) \leftarrow \text{Softmax}(\textbf{W}_{ctx}^\top h_{\hat{\theta}}(x))$
\State Compute KL divergence loss $\mathcal{L}_{KL} \leftarrow \textbf{KL}[p_{ctx}(x;\theta)\|p_{ctx}(x;\hat{\theta})]$
\State Compute classification loss $\mathcal{L}_{CE}(\textbf{W}_{f}^\top h_{\hat{\theta}}(x), y)$
\State $\mathcal{L} \leftarrow \mathcal{L}_{CE}+\alpha\mathcal{L}_{KL}$ \hfill
\State Update parameters $\hat{\theta}$, $\textbf{W}_{f}$ for minimizing $\mathcal{L}$
\\ 
\EndFor
\end{algorithmic}
\label{alg:carft}
\end{algorithm*}
\section{Method}

\subsection{Preliminaries} \label{sec:pre}
Vision-Language Pre-training (VLP) models like CLIP~\citep{radford2021learning} and ALIGN~\citep{jia2021scaling} consist of an image encoder $h_{\theta}(\cdot)$ and text encoder $g_{\phi}(\cdot)$ with dual form. Both of them are pre-trained by contrastive loss~\citep{hadsell2006dimensionality} to push the embedding of matched image-text pairs together while pushing those of mismatched pairs apart. After pre-training, we can get the visual representation $h_{\theta}(x) \in \mathbb{R}^{D}$ corresponding to input image $x$, $D$ is feature dimensions. 

\noindent\textbf{Zero-shot CLIP.} Consider a downstream classification task with image $x\in \mathcal{X}$ and label $y\in \mathcal{Y}$, $\lvert \mathcal{Y}\rvert=K$. Borrowing the pre-training objective of text-image matching, we can design text prompt to fit CLIP to zero-shot classification. Specifically, let $\mathcal{P}$ be a set of prompt templates such like ``a $\mathtt{[CONTEXT]}$ of $\mathtt{[CLASS]}$.'', where $\mathtt{[CLASS]}$, $\mathtt{[CONTEXT]}$ are the placeholder for a specific class name and context description respectively. $\mathcal{C}$ is a set of class names corresponding to each $y\in \mathcal{Y}$. All combinations of prompt templates and class names consist of text prompt set $\mathcal{T}=\mathcal{P}\times \mathcal{C}$, we feed them to text encoder to get $\textbf{W}=g_{\phi}(\mathcal{T}) \in \mathbb{R}^{D\times \lvert \mathcal{P}\rvert \times \lvert \mathcal{C}\rvert}$, where $\lvert \mathcal{C}\rvert=K$. The final classification weights can be calculated by: 
\begin{equation} \label{eq:obj_prompt}
    \textbf{W}_{cls} = \textbf{Norm}(\frac{\sum_{i=0}^{\lvert \mathcal{P}\rvert}\textbf{W}_{[:,i,:]}}{\lvert \mathcal{P}\rvert}) \in \mathbb{R}^{D \times K},
\end{equation}
where $\textbf{W}_{cls}$ is the text features corresponding to each class, and $\textbf{Norm}(\cdot)$ is the Frobenius norm. If not specified, both $\textbf{W}_{cls}$ and $h_{\theta}(x)$ have been normalized, such that we can compare $\textbf{W}_{cls}$ with image features $h_{\theta}(x)$, and use their cosine similarity $\textbf{W}_{cls}^\top h_{\theta}(x)$ as the zero-shot prediction.

\noindent\textbf{Fine-tuned CLIP.} A standard fine-tuning paradigm always adds a set of learnable parameters $\textbf{W}_{f}\in \mathbb{R}^{D\times K}$ for mapping pre-trained representation into label space of downstream task, where $\textbf{W}_{f}$ is usually randomly initialized. Since CLIP has its own zero-shot ability, a better choice is to use zero-shot classification weights $\textbf{W}_{cls}$ for initialization instead of random initialization~\citep{wortsman2022robust,wortsman2022model}. Both $\theta$ and $\textbf{W}_{f}$ are optimized by minimizing cross-entropy loss. Note that we only consider the end-to-end fine-tuning in this work, which usually brings best results when training data is sufficient.

\subsection{An Empirical Study of CLIP's Context-Awareness} \label{sec:context_study}
In this section, we empirically study the ability of CLIP models recognizing specific context, and how it changes during fine-tuning. To this end, we construct a context recognition task, whose aim is to classify four contexts, i.e. art painting, cartoon, photo, sketch, contained in PACS dataset~\citep{li2017deeper}. As suggested in Figure~\ref{fig:study}, the original CLIP $h_{\theta}$ and model $h_{\hat{\theta}}$ fine-tuned with category label are compared on this task. Since class names and context descriptions have been given in PACS, we follow the procedure of zero-shot CLIP in Section~\ref{sec:pre} to get prompt weights $\textbf{W}$. The zero-shot context prompt weights are then obtained by:
\begin{equation} \label{eq:ctx_prompt}
    \textbf{W}_{ctx} = \textbf{Norm}(\frac{\sum_{i=0}^{\lvert \mathcal{C}\rvert}\textbf{W}_{[:,:,i]}}{\lvert \mathcal{C}\rvert}) \in \mathbb{R}^{D \times \lvert \mathcal{P}\rvert},
\end{equation}
which is used for classifying contexts. Interestingly, we find a zero-shot CLIP model can achieved 82.5\% accuracy, exhibiting a good context-aware capability. But such ability is soon lost in fine-tuning stage, the context recognition accuracy of fine-tuned CLIP are fell into 26.7\%, close to a random guess. Above investigation shows that downstream fine-tuning obscures context-related features. As consequence, the fine-tuned CLIP models cannot take advantage of context-awareness for classifying images with unusual contexts, which results in the underperformance on OOD classification. 

\subsection{Context-Aware Robust Fine-tuning}
We introduce the proposed Context-Aware Robust Fine-tuning (CAR-FT) in this section. Based on empirical analysis in Section~\ref{sec:context_study}, we are motivated to retain the context-awareness of CLIP models during fine-tuning. Our main idea lies in using zero-shot CLIP models to induce the context-related information in guiding with fine-tuning process. Such that a context-aware feature can be learnt on downstream tasks for achieving both advanced ID and OOD performance. Specifically, CAR-FT uses context prompt weights $\textbf{W}_{ctx}$ to get the reference
distribution of context in the input image:
\begin{equation}
    p_{ctx}(x;\theta) = \text{Softmax}(\textbf{W}_{ctx}^\top h_{\theta}(x))\in \mathbb{R}^{\lvert \mathcal{P}\rvert},
\end{equation}
where $\theta$ is the image encoder weights of zero-shot CLIP. At start of fine-tuning stage, we duplicate a new copy of $\theta$ as $\hat{\theta}$. We assume fine-tuned CLIP model is parameterized by $\hat{\theta}$, and $\textbf{W}_{ctx}$ is shared between pre-trained and fine-tuned CLIP. The induced context distribution of fine-tuned models can be:
\begin{equation}
    p_{ctx}(x;\hat{\theta}) = \text{Softmax}(\textbf{W}_{ctx}^\top h_{\hat{\theta}}(x))\in \mathbb{R}^{\lvert \mathcal{P}\rvert}.
\end{equation}
To regularize the fine-tuned visual representation i.e. $h_{\hat{\theta}}(x)$ to encode effective context information, we make the predicted context distribution $p_{ctx}(x;\hat{\theta})$ closer to $p_{ctx}(x;\theta)$. This regularization term is realized by minimizing the Kullback-Leibler Divergence (KLD):
\begin{equation}
    \mathcal{L}_{KL}=\textbf{KL}[p_{ctx}(x;\theta)\|p_{ctx}(x;\hat{\theta})].
\end{equation}
Another objective of CAR-FT is the regular downstream classification loss. The overall loss of our CAR-FT is:
\begin{equation} \label{eq:overall_loss}
    \mathcal{L}= \mathcal{L}_{CE}(\textbf{W}_{f}^\top h_{\hat{\theta}}(x), y) + \alpha \mathcal{L}_{KL}.
\end{equation}
$\alpha$ is a factor to trade-off the impact of $\mathcal{L}_{KL}$ term. In this work we set $\alpha=1$ empirically. It should also be denoted that only $\hat{\theta}$ and $\textbf{W}_{f}$ is updated during fine-tuning and all other parameters are frozen. The 
detailed procedure of our CAR-FT is summarized in Algorithm~\ref{alg:carft}.

\section{Experiments}
We demonstrate the robustness of Context-Aware Robust Fine-tuning (CAR-FT) against distribution shifts by evaluating on large-scale ImageNet classification and DomainBed which consists of five Domain Generalization (DG) tasks. We further show CAR-FT gains accuracy improvement on downstream tasks in Section~\ref{sec:transfer}. Finally thorough ablation experiments are conducted to study the impact of hyper-parameters, model scales and so on. If not specified, for all experiments we adopt ViT-B/16 as default backbone. 

\subsection{Robustness to Distribution Shifts}

\subsubsection{ImageNet Classification} \label{sec:imagenet_exp}

\noindent\textbf{Benchmarks.} We use five OOD testsets on ImageNet classification task. Each of them represents a type of OOD scenario where the classifier is prone to make mistakes. IN-V2~\citep{recht2019imagenet} is a new ImageNet test set with distribution shift; IN-R~\citep{hendrycks2021many} collects online images with artificial creation, e.g., cartoons, graphics, video game renditions, etc; IN-Sketch~\citep{wang2019learning} contains sketches instead of natural images; ObjectNet~\citep{barbu2019objectnet} places ImageNet objects in hard contexts, e.g. unusual backgrounds, rotations or imaging viewpoints; IN-A~\citep{hendrycks2021natural} consists of images misclassified by a ResNet-50~\citep{he2016deep}. We report the top-1 accuracy on all above datasets. 

\noindent\textbf{Implementation.} For training large-scale ImageNet dataset, we adopt AdamW optimizer~\citep{loshchilov2018decoupled} with initial learning rate of 3e-5, weight decay of 0.1 and a cosine annealing decay schedule. Only random resized crop is used for data augmentation. All models are trained 10 epochs with batch size of 512.

\begin{table*}[t]\footnotesize
\begin{center}
\caption{Top@1 accuracy of compared methods on ImageNet and its derived distribution shifts. Avg. shifts presents the mean accuracy among five distribution shifts. We use ViT-B/16 as basic backbone.} \label{tab:imagenet}
\begin{tabular}{lc|ccccc|cc}
\toprule
 & & \multicolumn{5}{c|}{Distribution shifts} & Avg. & Avg. \\
 & IN & IN-V2 & IN-R & IN-Sketch & ObjectNet & IN-A & shifts & all \\
\midrule
CLIP Zero-shot & 68.3 & 61.9 & 77.6 & 48.3 & 29.8 & 50.1 & 53.5 & 56.0 \\
\texttt{Fine-tuning Only Methods}  & & & & & & & \\
\quad  FT & 81.0 & 70.9 & 54.7 & 42.1 & 26.6 & 31.3 & 45.1 & 51.1 \\
\quad  TP-FT & 81.2 & 70.7 & 65.0 & 44.9 & 27.4 & 35.3 & 48.7 & 54.1 \\
\quad  LP-FT & 81.7 & 71.6 & 72.9 & 48.4 & 28.2 & 49.1 & 54.0 & 58.7 \\
\rowcolor{tabhighlight}
\quad  CAR-FT (Ours) & \textbf{83.3} & \textbf{74.0} & \textbf{75.4} & \textbf{53.0} & \textbf{32.6} & \textbf{49.5} & \textbf{56.9} & \textbf{61.3}\\
\midrule
\texttt{Combined with Weight-Space Ensemble}  & & & & & & & \\
\quad WiSE-FT (opt. $\alpha$) & 81.7 & 72.7 & 78.8 & 53.2 & 33.4 & 52.2 & 58.1 & 62.0 \\
\rowcolor{tabhighlight}
\quad + CAR-FT & \textbf{82.1} & \textbf{73.3} & \textbf{79.2} & \textbf{54.5} & \textbf{33.8} & \textbf{53.6} & \textbf{58.9} & \textbf{62.8} \\
\midrule
\quad Greedy Model Soups & 84.8 & 75.1 & 74.1 & 53.8 & 30.3 & 48.1 & 56.3 & 61.0 \\
\rowcolor{tabhighlight}
\quad + CAR-FT & \textbf{85.0} & \textbf{75.8} & \textbf{74.4} & \textbf{54.6} & \textbf{31.6} & \textbf{48.9} & \textbf{57.1} & \textbf{61.7} \\
\midrule
\quad Uniform Model Soups & 83.4 & 74.7 & 76.8 & 54.6 & 31.5 & 50.0 & 57.5 & 61.8 \\
\rowcolor{tabhighlight}
\quad + CAR-FT & \textbf{83.9} & \textbf{75.1} & \textbf{77.3} & \textbf{55.5} & \textbf{32.0} & \textbf{51.1} & \textbf{58.2} & \textbf{62.5}\\

\bottomrule
\end{tabular}
\end{center}
\end{table*}

\noindent\textbf{Results.} We first compare our CAR-FT with previous fine-tuning methods. Among them, FT, TP-FT~\citep{li2022elevater,wortsman2022robust,wortsman2022model}, LP-FT~\citep{kumar2021fine} adopts random weights, text prompt weights and linear probing weights for initialization of the classification head respectively. It can be shown that linear probing weights is better for initializing classification layer for downstream fine-tuning. However, to get linear probing weights, LP-FT must spend extra round of training. By contrast, our CAR-FT, which uses text prompt weights for initialization, has less training cost. Meanwhile, compared with LP-FT, CAR-FT achieves +1.6\% and 2.9\% improvement on ID and OOD accuracy. It suggests that preserving context-awareness during fine-tuning does help for OOD classification. 

To demonstrate the versatility of our method, we combine CAR-FT with weight-space ensemble methods to further robustify the model. Both WiSE-FT~\citep{wortsman2022robust} and ModelSoups~\citep{wortsman2022model} take advantage from the shared training trajectory among models fine-tuned with same pre-trained weights, and use weight-space ensemble to integrate the power of multiple models. Our CAR-FT can be applied by simply modifying preceding fine-tuning stage of these method. For WiSE-FT, we use the optimal interpolation weight for comparison. The results are shown in Table~\ref{tab:imagenet}. An interesting phenomenon is that the ID accuracy of CAT-FT drops from 83.3\% to 82.1\% after combining with WiSE-FT. This finding is contrary to WISE-FT paper, where weight-space ensemble always helps improving accuracy. We think CAR-FT has learnt useful knowledge from the zero-shot CLIP models via KLD loss. After that, assembling with a low accurate zero-shot model has no complementary promotion but even neutralizes the original performance. However, although accuracy is decreased, the robustness of CAR-FT under distribution shifts is enhanced with weight-space ensemble. Our methods can yield +0.8\% averaged OOD accuracy than WiSE-FT. ModelSoups fine-tune 72 models with various combinations of hyper-parameters, and conduct weight-space ensemble of them with uniform or greedy policy. We test CAR-FT augmented with uniform and greedy soups. Our CAR-FT can further obtain +0.2\%, +0.5\% ID and +0.8\%, +0.7\% OOD accuracy of greedy soup and uniform soup respectively.

\subsubsection{Domain Generalization}\label{sec:dg}
\noindent\textbf{Benchmarks.} We present our results on five benchmark datasets included in DomainBed~\citep{gulrajani2020search}: PACS~\citep{li2017deeper} (4 domains, 7 classes, and 9, 991 images), VLCS~\citep{torralba2011unbiased} (4 domains, 5 classes, and 10, 729 images), OfficeHome~\citep{venkateswara2017deep} (4 domains, 65 classes, and 15, 588 images), TerraIncognita~\citep{beery2018recognition} (4 domains, 10 classes, and 24, 788 images), and DomainNet~\citep{peng2019moment} (6 domains, 345 classes, and 586, 575 images). The leave-one-out evaluation protocol is adopted, where we iteratively choose a single domain as test domain and use other domains for training. The averaged accuracy on all chosen test domains is reported as the general performance on each dataset. We repeat experiment of CAR-FT three times for suppressing the fluctuation of results. 

\begin{table*}[t]\footnotesize
\begin{center}
\caption{Comparison with domain generalization methods on DomainBed. The accuracy reported are averaged on three trials. }\label{tab:dg}
\begin{tabular}{l|ccccc|c}
\toprule
  & PACS & VLCS & OfficeHome & TerraInc & DomainNet & Avg. \\
\midrule
\multicolumn{7}{l}{\texttt{Previous SOTA using RegNetY-16GF}} \\
MIRO~\citep{cha2022domain} & \textbf{97.4} & 79.9 & 80.4 & 58.9 & 53.8 & 74.1 \\
EoA~\citep{arpit2021ensemble}  & 95.8 & 81.1 & 83.9 & 61.1 & 60.9 & 76.6 \\
MIRO+SWAD~\citep{cha2022domain,cha2021swad} & 96.8 & 81.7 & 83.3 & \textbf{64.3} & 60.7 & 77.3 \\
\midrule

\multicolumn{7}{l}{\texttt{ViT-B/16 Pre-trained by CLIP}} \\
ERM & 93.7 & 82.7 & 78.5 & 52.3 & 53.8 & 72.2 \\
MIRO~\citep{cha2022domain}  & 95.6 & 82.2 & 82.5 & 54.3 & 54.0 & 73.7 \\
DPL~\citep{zhang2022domain} & 97.3 & 84.3 & 84.2 & 52.6 & 56.7 & 75.0 \\
\rowcolor{tabhighlight}
CAR-FT (Ours)  & 96.8 & \textbf{85.5} & \textbf{85.7} & 61.9 & \textbf{62.5} & \textbf{78.5} \\
\bottomrule
\end{tabular}
\end{center}
\end{table*}
\begin{table*}[t]\footnotesize
\begin{center}
\caption{Performance on nine downstream tasks using different fine-tuning methods. The reported accuracy is averaged on three trials. }\label{tab:downstream}
\begin{tabular}{l|ccccccccc|c}
\toprule
 Methods & ImageNet & Flowers & Aircraft & Pets & CIFAR10 & CIFAR100 & Cars & DTD & SUN397 & Avg. \\
\midrule
FT & 81.0 & 93.7 & 53.0 & \textbf{93.4} & 97.6 & 85.6 & 88.1 & 75.8 & 71.5 & 82.2\\
TP-FT & 81.2 & 93.1 & 51.6 & 90.6 & 97.7 & 89.5 & 86.6 & 77.4 & 74.8 & 82.5 \\
LP-FT & 81.7 & \textbf{95.0} & 53.3 & 93.2 & 97.8 & 85.4 & 88.5 & 75.4 & 71.6 & 82.4 \\
WiSE-FT (opt. $\alpha$) & 82.5 & 93.6 & 53.2 & 90.7 & 98.2 & 90.6 & 88.3 & 78.5 & 78.6 & 83.8\\
\rowcolor{tabhighlight}
CAR-FT (Ours) & \textbf{83.3} & 94.4 & \textbf{54.1} & 92.5 & \textbf{98.7} & \textbf{91.1} & \textbf{88.9} & \textbf{79.0} & \textbf{78.9} & \textbf{84.5}\\

\bottomrule
\end{tabular}
\end{center}
\end{table*}

\noindent\textbf{Implementation. } Our CAR-FT requires designed context prompts. However, for Domain Generalization (DG) datasets, there is no off-the-shelf description for each class or context. So we construct the prompts ourselves. For PACS, OfficeHome and DomainNet, since domain names have provided, we use a template of ``a $\mathtt{[CONTEXT]}$ of $\mathtt{[CLASS]}$.'', where $\mathtt{[CONTEXT]}$ is the domain name. Take PACS as an example, $\mathtt{[CONTEXT]}$ is the one of \{art painting, cartoon, photo, sketch\}. For TerraIncognita and VLCS, even the domain names are not available. We directly use the prompt templates of ImageNet in CLIP paper~\citep{radford2021learning}. The detailed text prompts of each dataset can be referred to Supplementary. Next we present the training configuration. The models in domain generalization experiments are optimized by AdamW with learning rate of 5e-6 and weight decay of 0.1. For data augmentation, simple random resize crop and random horizontal filp are used. The other hyper-parameters, such as batch size, dropout rate, and training steps, we keep consistent with the default configuration in DomainBed. 

\noindent\textbf{Results.} We make fair comparison with ERM, MIRO~\citep{cha2022domain}, and DPL~\citep{zhang2022domain}. All of them use ViT-B/16 pre-trained by CLIP on 4B image-text pairs. Benefit from large-scale pre-training on massive data, simple ERM baseline can obtain 72.2\% averaged accuracy on five datasets. The recent proposed DG algorithms MIRO and DPL yield +1.5\% and +2.8\% average improvement respectively. By contrast, our CAR-FT surpasses all other methods and achieves largest +6.3\% improvement. We also compare CAR-FT with previously reported state-of-the-art of DG task. The best reported performance is MIRO with SWAD~\citep{cha2021swad} using a RegNetY-16GF~\citep{radosavovic2020designing} pre-trained on 3.6B images. Our CAR-FT builds the new state-of-the-art with the leading of +1.2\%. 

\subsection{Accuracy on Downstream Tasks} \label{sec:transfer}

\noindent\textbf{Benchmarks.} Beyond robustness to distribution shift, we also select 9 image classification datasets used in downstream transfer exmperiment of CLIP: ImageNet~\citep{deng2009imagenet}, CIFAR10, CIFAR100, FGVCAircraft~\citep{maji2013fine}, OxfordPets~\citep{parkhi2012cats}, StanfordCars~\citep{krause20133d}, Flowers102~\citep{nilsback2008automated}, SUN397~\citep{xiao2016sun} and DTD~\citep{cimpoi2014describing} (see Supplementary for their statistics). These benchmarks cover a diverse tasks of classification on textures, image details and generic objects. We use them for comprehensive evaluation of compared fine-tuning methods. 

\noindent\textbf{Implementation.} For each downstream dataset, we use the corresponding text prompt in CLIP repository\footnote{\url{https://github.com/openai/CLIP/blob/main/data/prompts.md}}. The model is end-to-end fine-tuned. Other training settings are consistent with ImageNet experiment in Section~\ref{sec:imagenet_exp}. 

\noindent\textbf{Results.} We fine-tune CLIP models with various method and report the top@1 accuracy on downstream tasks. Baseline methods in Section~\ref{sec:imagenet_exp} are compared except for ModelSoups, which assembles a large number of models and may bring unfairness of comparison. Since we have shown previously that WiSE-FT cannot improve the ID accuracy of CAR-FT models, in this experiment, we only use plain CAR-FT without augmenting of weight-space ensemble methods. The results are shown in Table~\ref{tab:downstream}. Generally, text prompts can be regarded as good prior to initialize the classification head. However, our experiment presents TP-FT falls behind of FT on several datasets e.g. Flowers102, OxfordPets and StanfordCars, which demonstrates that fine-tuning a randomly initialized head is even better than initialization with text prompts in some cases. Besides, for other methods, both WiSE-FT and our CAR-FT improve the baseline of TP-FT consistently. Our CAR-FT achieves the best averaged accuracy, which surpasses WiSE-FT by 0.7\%. This experiment suggests learning a context-aware feature can also benefit the performance on in-domain data. 

\begin{table}[t]\footnotesize
\begin{center}
\caption{The effect of CAR-FT on different CLIP backbones. } \label{tab:backbone}
\begin{tabular}{lc|c}
\toprule
 & ImageNet & Avg. shifts \\
\midrule
\texttt{ResNet50} &  &  \\
\quad  Zero-shot & 59.6 & 38.5 \\
\quad  TP-FT & 76.1 & 37.6 \\
\quad  LP-FT & 76.2 & 39.6 \\
\quad  WiSE-FT & 76.2 & 42.2 \\
\rowcolor{tabhighlight}
\quad  CAR-FT & \textbf{76.7} & 42.9 \\
\rowcolor{tabhighlight}
\quad  CAR-FT+WiSE-FT & 76.3 & \textbf{43.3} \\
\midrule
\texttt{ViT-L/14} & & \\
\quad  Zero-shot & 75.5 & 65.1 \\
\quad  TP-FT & 85.0 & 59.9 \\
\quad  LP-FT & 85.3 & 66.2 \\
\quad  WiSE-FT & 86.0 & 68.8 \\
\rowcolor{tabhighlight}
\quad  CAR-FT & \textbf{87.1} & 67.8 \\
\rowcolor{tabhighlight}
\quad  CAR-FT+WiSE-FT & 86.6 & \textbf{69.7} \\
\midrule
\texttt{ViT-L/14@336px} & & \\
\quad  Zero-shot & 76.2 & 67.3 \\
\quad  TP-FT & 86.2 & 64.2 \\
\quad  LP-FT & 86.3 & 68.0 \\
\quad  WiSE-FT & 87.1 & 71.7 \\
\rowcolor{tabhighlight}
\quad  CAR-FT & \textbf{87.7} & 70.2 \\
\rowcolor{tabhighlight}
\quad  CAR-FT+WiSE-FT & 87.3 & \textbf{72.1} \\
\midrule
\texttt{ViT-L/14 (using LAION-2B)} & & \\
\quad  Zero-shot & 75.2 & 62.0 \\
\quad  TP-FT & 84.3 & 59.4 \\
\quad  LP-FT & 84.4 & 62.8 \\
\quad  WiSE-FT & 85.3 & 66.0 \\
\rowcolor{tabhighlight}
\quad  CAR-FT & \textbf{86.1} & 66.0 \\
\rowcolor{tabhighlight}
\quad  CAR-FT+WiSE-FT & 85.6 & \textbf{67.1} \\
\midrule
\texttt{ViT-H/14 (using LAION-2B)} & & \\
\quad  Zero-shot & 78.0 & 65.2 \\
\quad  TP-FT & 84.7 & 58.7 \\
\quad  LP-FT & 84.9 & 66.1 \\
\quad  WiSE-FT & 86.0 & 68.2 \\
\rowcolor{tabhighlight}
\quad  CAR-FT & \textbf{86.6} & 69.7 \\
\rowcolor{tabhighlight}
\quad  CAR-FT+WiSE-FT & 86.3 & \textbf{69.9} \\
\bottomrule
\end{tabular}
\end{center}
\end{table}

\subsection{Ablation Study}

\begin{figure}[t]
    \centering
    \includegraphics[width=0.4\textwidth]{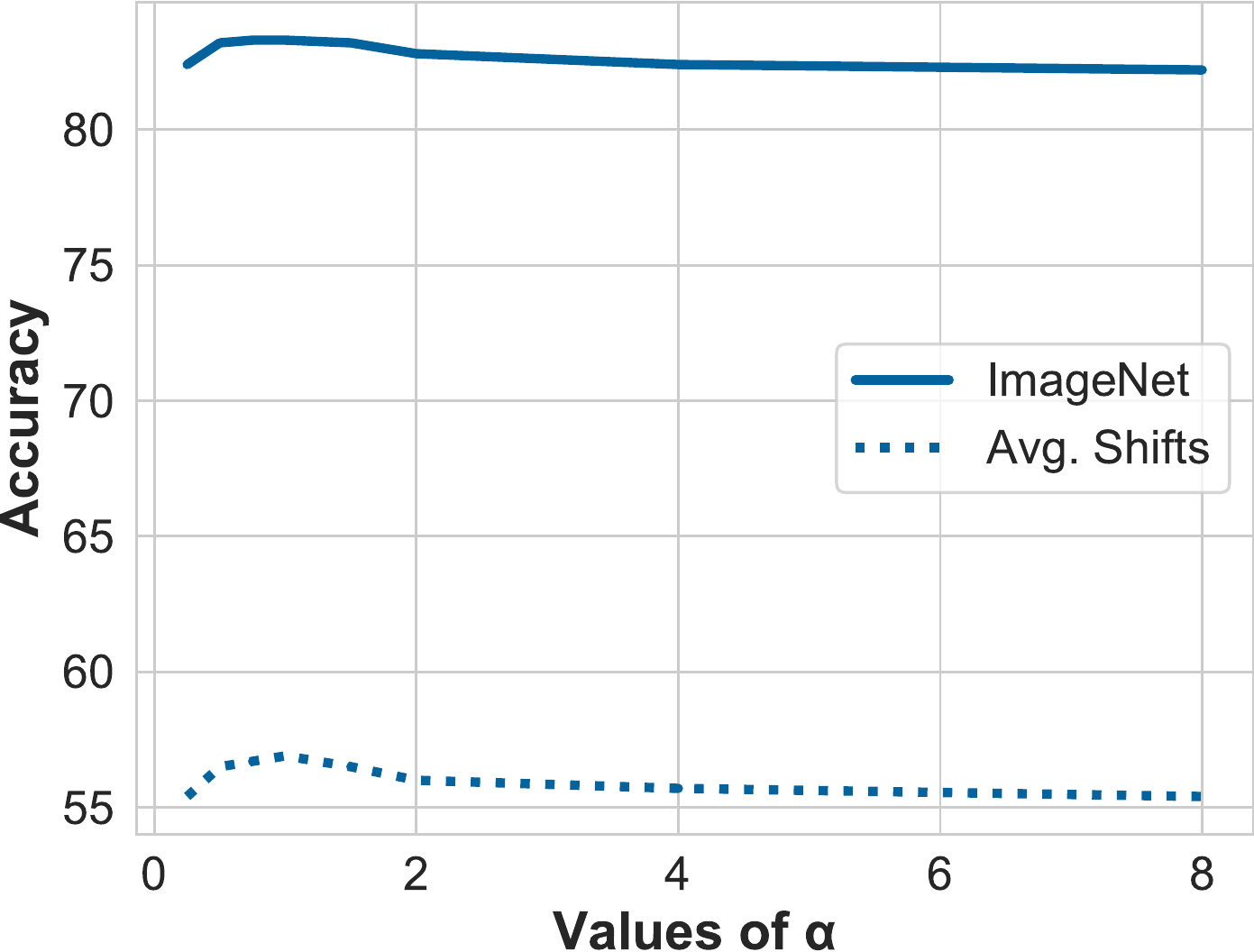}
    \caption{The performance of our CAR-FT with different $\alpha$ on ImageNet. The solid line indicates the validation accuracy, and dashed line presents the averaged accuracy under distribution shifts. }
    \label{fig:ablation_alpha}
\end{figure}

\noindent\textbf{Trade-off Between Loss Terms.} We set a hyper-parameter $\alpha$ in Equation~\ref{eq:overall_loss} to balance the effect of two loss terms. To study the sensibility of our method w.r.t $\alpha$, we adopt multiple values of $\alpha$ for running CAR-FT on ImageNet. The results are shown in Figure~\ref{fig:ablation_alpha}. The best performance appears at around $\alpha=1$, which is the empirical value used in this paper. Besides, we find both the ID and OOD accuracy have roughly the same trend with change of $\alpha$. And they become worse along with the smaller or larger value of $\alpha$. The figure also suggests CAR-FT is insensitive with larger values of $\alpha$. The performance decreases slowly after $\alpha$ is larger than 2. This phenomenon also indicates KLD loss term is the key component for the enhancement. 

\begin{figure*}[t]
    \centering
    \includegraphics[width=1.0\textwidth]{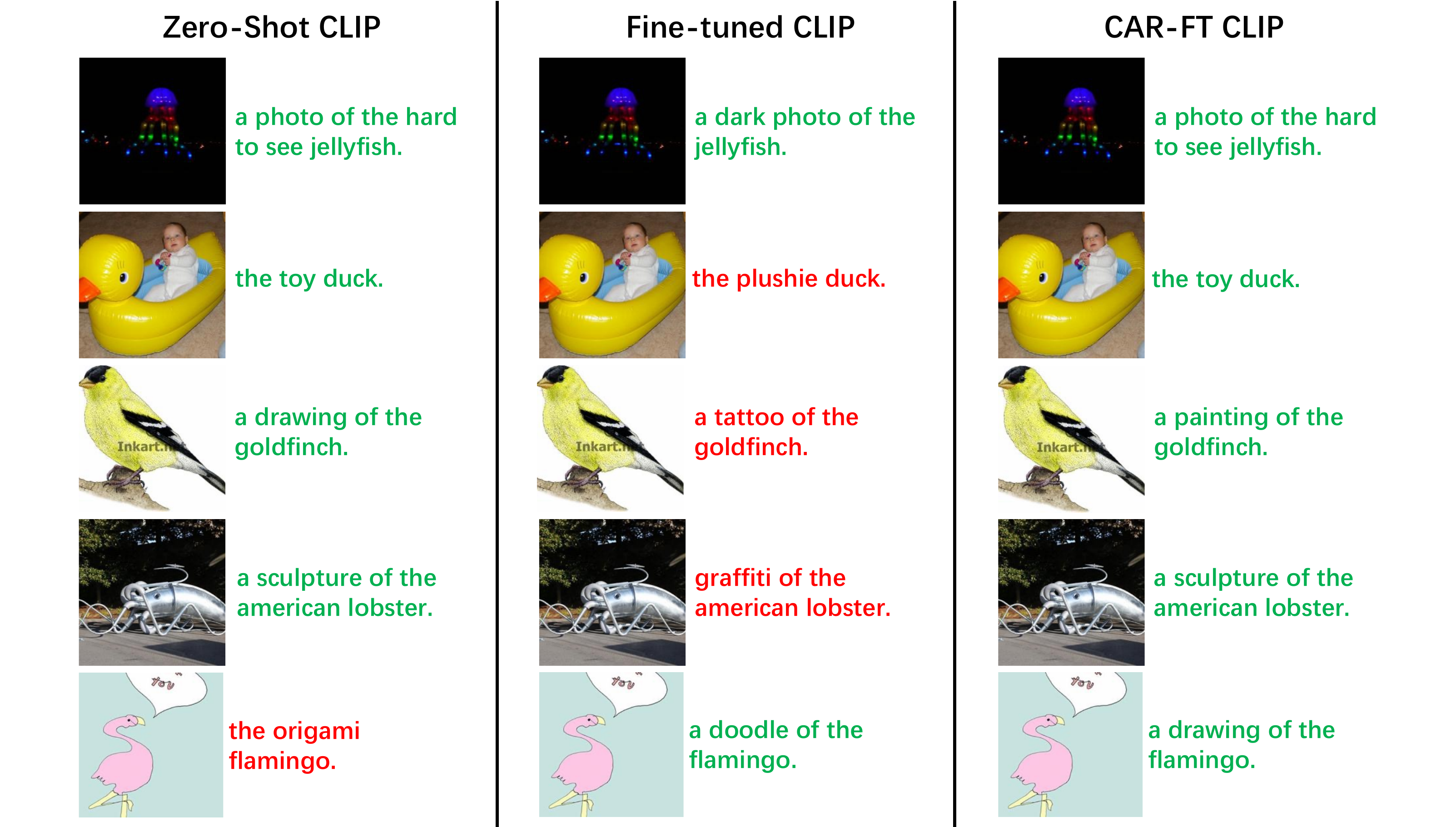}
    \caption{Visualization of top@1 model prediction on context prompt. The text in red presents incorrect prediction. }
    \label{fig:viscontext}
\end{figure*}

\noindent\textbf{Different Backbones.} We conduct ablation on backbones to learn how model scales or architectures affect the performance of our CAR-FT in Table~\ref{tab:backbone}. For simplicity, we omit the detailed accuracy on each OOD dataset and directly show the averaged accuracy under 5 distribution shifts. ResNet50 and ViT are adopted as the typical convolution-based and transformer-based networks respectively. We adopt the official weights of CLIP for ResNet50 and ViT-L/14, ViT-L/14@336px. For more types of backbones, we additionally use the weights of ViT-L/14 and ViT-H/14 opened in OpenCLIP~\citep{ilharco_gabriel_2021_5143773}, which are pre-trained on LAION-2B~\citep{schuhmannlaion}. Our CAR-FT outperforms other methods across multiple model scales and architectures. Especially, a plain CAR-FT without weight-space ensemble can even surpass WiSE-FT on ResNet50 and ViT-H/14. By sacrificing little accuracy on ImageNet, the robustness against distribution shifts of CAR-FT can be further improved by WiSE-FT. However, such trade-off results are at least better than WiST-FT, with both higher ID and OOD accuracy.  

\begin{table}[t]
\begin{center}
\caption{CAR-FT with different types or amounts of text prompts. } \label{tab:template}
\begin{tabular}{c|c|c|c}
\toprule
Template & Template & \multirow{2}{*}{ImageNet} & Avg. \\
Types & Num &  & shifts \\
\midrule
CLIP & 80 & 83.3 & 56.9 \\
CLIP & 7 & 83.2 & 56.5 \\
Searched & 7 & 83.2 & 56.9 \\
\bottomrule
\end{tabular}
\end{center}
\end{table}

\begin{table}[t]\footnotesize
\begin{center}
\caption{Performance of CAR-FT on other CLIP variants.} \label{tab:beyond}
\begin{tabular}{lc|c}
\toprule
 & ImageNet & Avg. shifts \\
\midrule
\texttt{ViT-B/32 by DeCLIP-88M} &  &  \\
\quad  Zero-shot & 66.2 & 43.0 \\
\quad  TP-FT & 74.2 & 38.9 \\
\quad  LP-FT & 74.0 & 42.8 \\
\rowcolor{tabhighlight}
\quad  CAR-FT & \textbf{74.3} & \textbf{43.7} \\
\midrule
\texttt{ViT-B/32 by FILIP-YFCC15M} & & \\
\quad  Zero-shot & 39.5 & 19.6 \\
\quad  TP-FT & 59.4 & 24.2 \\
\quad  LP-FT & 59.3 & 24.1 \\
\rowcolor{tabhighlight}
\quad  CAR-FT & \textbf{59.4} & \textbf{24.4} \\
\midrule
\texttt{ViT-B/32 by SLIP-YFCC15M} & & \\
\quad  Zero-shot & 34.3 & 15.5 \\
\quad  TP-FT & 56.0 & 20.7 \\
\quad  LP-FT & 56.1 & 20.7 \\
\rowcolor{tabhighlight}
\quad  CAR-FT & \textbf{56.1} & \textbf{21.0} \\
\bottomrule
\end{tabular}
\end{center}
\end{table}

\noindent\textbf{Impact of Context Prompts.} In our CAR-FT, prompts provide language supervision of context. This supervision is influenced by types or quantity of used prompts. In general understanding, the finer granularity and greater quantity of used prompts, the more types of image contexts can be perceived by CAR-FT. However, we find a few of concise context prompts are sufficient for achieving desirable results of CAR-FT. This finding is validated by our ablation experiment comparing CAR-FT models by various types and quantity of used prompts. Specifically, we adopt official CLIP templates~\citep{radford2021learning}, and change the size of prompts by randomly sampling a subset from original template set. The subset size of 7 is compared. Results in Table~\ref{tab:template} suggest only using 7 context prompts gets lower ID and OOD performance. To compare different types of context prompt, we use the searched 7 templates\footnote{\url{https://github.com/openai/CLIP/blob/main/notebooks/Prompt_Engineering_for_ImageNet.ipynb}}. After using better organized prompts, the OOD accuracy ups to 56.9\%, which is same performed with baseline model. 

\begin{figure*}[t]
    \centering
    \includegraphics[width=0.95\textwidth]{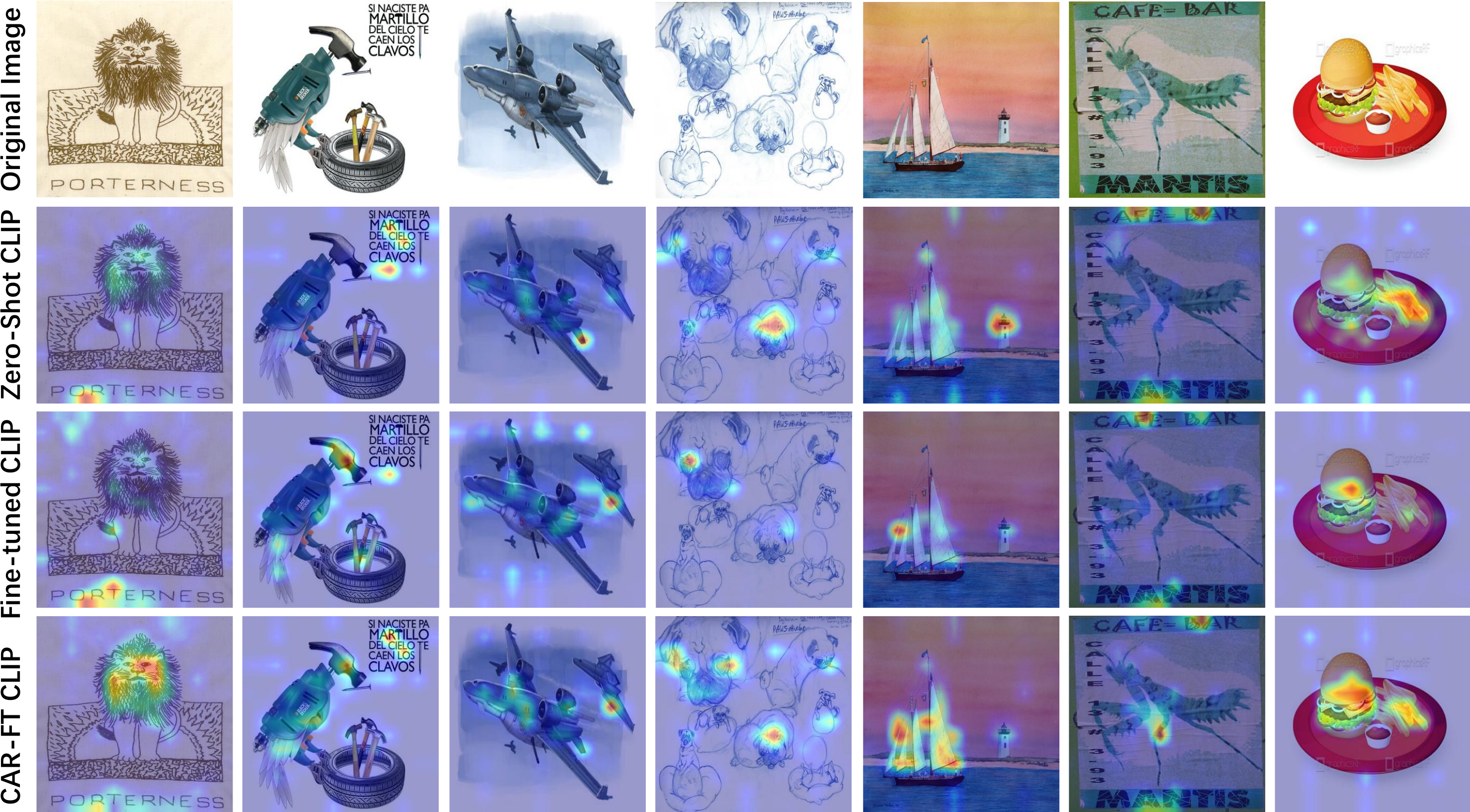}
    \caption{Visualized attention of CLIP models with zero-shot, traditional fine-tuning and our CAR-FT.}
    \label{fig:attention}
\end{figure*}

\begin{figure*}[t]
    \centering
    \includegraphics[width=1.0\textwidth]{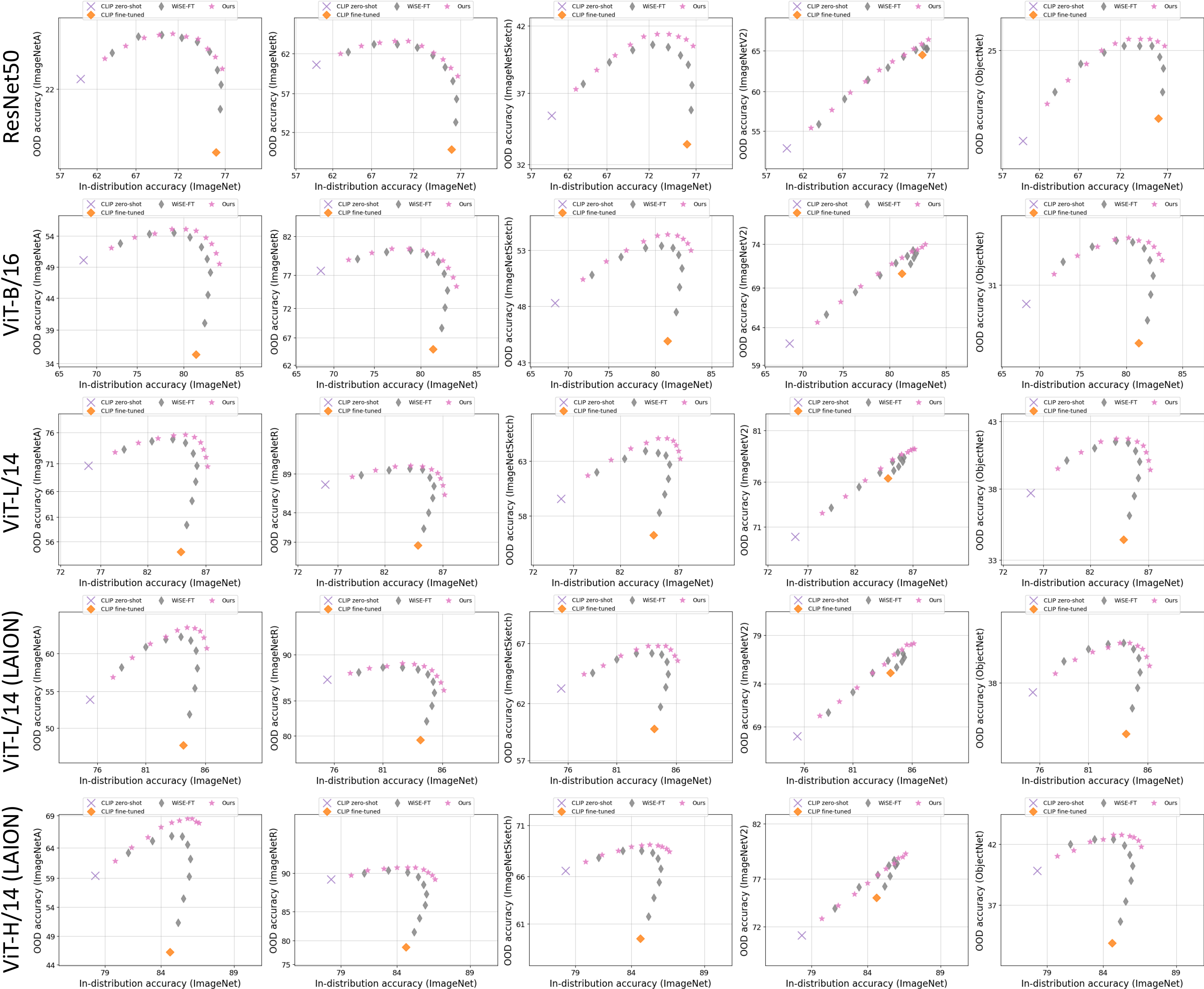}
    \caption{More results of interpolated weights using WiSE-FT. We compare original WiSE-FT with our CAR-FT based WiSE-FT.}
    \label{fig:wiseft}
\end{figure*}

\noindent\textbf{Beyond CLIP.} Table~\ref{tab:beyond} illustrates if CAR-FT is still helpful on other CLIP variants such as DeCLIP~\citep{li2021supervision}, SLIP~\citep{mu2022slip} or FILIP~\citep{yao2021filip}. Among them, DeCLIP is pre-trained on private 88M dataset, and others are pre-trained on YFCC15M, a filtered version of YFCC100M~\citep{thomee2016yfcc100m}. We discover that the effect of CAR-FT relies on the zero-shot classification capability. For zero-shot models pre-trained on smaller datasets with low performance, e.g. FILIP and SLIP, CAR-FT can only obtain +0.3\% improvement on OOD accuracy. It is reasonable that a poorly performed zero-shot model always cannot provide an accurate prediction of context distribution, which limits the promotion of our CAR-FT. 

\section{Discussion}
We further analyze the phenomena observed from CAR-FT and give some visualizations for deeper understanding about how CAR-FT works. 

\noindent\textbf{Can CAR-FT recognize image contexts? } Since the motivation of CAR-FT is to reduce the corruption of context-aware features during fine-tuning, it is necessary to discuss if CAR-FT models truly recognize image context successfully. We run the empirical experiment in Section~\ref{sec:context_study} to validate the context-awareness of our CAR-FT. We keep the other setting fixed and only replace the image encoder with the one fine-tuned by CAR-FT. The experimental result shows CAR-FT has 83.5\% zero-shot accuracy in recognizing contexts of PACS dataset, which is even 1\% higher than the original CLIP model. It reflects our method is effective in learning context-aware features. We additionally visualize the top@1 context prediction of CLIP models on ImageNet classification. We choose test samples in  ImageNet-R dataset and present the best matched context prompt in Figure~\ref{fig:viscontext}. It should be denoted that for single image, there will be more than one prompts matching its contexts. For example, in the first row of Figure~\ref{fig:viscontext}, both the description of ``a photo of the hard to see jellyfish.'' and ``a dark photo of the jellyfish.'' is reasonable in human cognition. We judge the correctness of a context prediction from a subjective perspective. For the zero-shot CLIP model, its most predictions are aligned with image context, conforming with human understanding. However, after fine-tuning, the model turns context-unaware and produces wrong text prompt prediction. Our CAR-FT can fix these mistakes and even output more accurate context predictions. The visualization reveals the strong ability of CAR-FT in context understanding vividly. Additionally, we use interpreting method of ~\cite{chefer2021generic} to visualize the attention of CAR-FT models. In Figure~\ref{fig:attention}, we sample some images in ImageNet-R with unusual context, and compare the attention on them using zero-shot, fine-tuned, and CAR-FT CLIP models. Our CAR-FT exhibits stronger interpretability, and focuses on the important objects even under uncommon context. 

\noindent\textbf{Other variants of CAR-FT.} There are multiple variants of CAR-FT in different ways of context prompt construction or context distribution divergence measuring. For example, Equation~\ref{eq:ctx_prompt} calculates $\textbf{W}_{ctx}$ by summing up context prompt weights of all categories. However, a subjectively more proper way is to choose the corresponding context prompt of 
the ground-truth category for $\textbf{W}_{ctx}$. We experiment with these two types of $\textbf{W}_{ctx}$ and find they have similar performance. For objective function, this paper uses KLD to measure the discrepancy of context distribution, actually there are more alternative metrics such as Maximum Mean Discrepancy (MMD) or Wasserstein distance. When we replace KLD with these advanced metrics, it still brings no better performance. Therefore, a simple KLD term is minimized in our CAR-FT.  

\section{Conclusion}

In this paper, we propose CAR-FT which alleviates the loss of context-aware ability of zero-shot models during fine-tuning. By inheriting context-awareness into downstream tasks, CAR-FT can build fine-tuned models with enhanced accuracy and robustness. Our work relies on image-language pre-trained models and task-related text prompts. Such strict preconditions make CAR-FT may lose its applicability on models pre-trained by only images or tasks without designed text prompts. How to apply the idea of context-aware fine-tuning on vision-only pre-trained models and remove the prerequisite of text prompts will be left as the future work.

\backmatter

\bmhead{Supplementary information} 

We provide the supplementary information about details of used datasets and prompts, and more results of CAR-FT combined with WiSE-FT.

\noindent\textbf{Datasets Details.} The detailed statistics of the nine downstream datasets, as well as the
five OOD testsets of ImageNet, are shown in Table~\ref{tab:datasets_details}.

\begin{table}[t]\footnotesize
\begin{center}
\caption{Details of the datasets used in our paper. For domain generalization benchmarks, we have introduced them in Section~\ref{sec:dg}.} \label{tab:datasets_details}
\begin{tabular}{lrrr}
\toprule
Dataset & Classes & Train & Test \\
\midrule
ImageNet & 1000 & 1,281,167 & 50,000 \\
CIFAR10 & 10 & 50,000 & 10,000 \\
CIFAR100 & 100 & 50,000 & 10,000 \\
FGVCAircraft & 100 & 6,667 & 3,333 \\
OxfordPets & 37 & 3,680 & 3,669 \\
StanfordCars & 196 & 8,144 & 8,041 \\
OxfordFlowers102 & 102 & 2,040 & 6,149 \\
SUN397 & 397 & 19,850 & 19,850 \\
Describable Textures & 47 & 3,760 & 1,880 \\
\midrule
ImageNet-V2 & 1,000 & N/A & 10,000 \\
ImageNet-R & 200 & N/A & 30,000 \\
ImageNet-Sketch & 1,000 & N/A & 50,889 \\
ObjectNet & 313 & N/A & 50,000 \\
ImageNet-A & 200 & N/A & 7,500 \\
\bottomrule
\end{tabular}
\end{center}
\end{table}

\noindent\textbf{Prompts Details.} Following is the prompt template used for PACS, OfficeHome and DomainNet in domain generalization experiments. For TerraIncognita and VLCS, we directly use the templates of ImageNet. 

\begin{lstlisting}[caption=Templates for PACS]
templates = ['a art painting of [CLASS].',
              'a cartoon [CLASS].',
              'a photo of [CLASS].',
              'a sketch of [CLASS].']
\end{lstlisting}

\begin{lstlisting}[caption=Templates for OfficeHome]
templates = ['a art painting of [CLASS].',
              'a clipart of [CLASS].',
              'a product of [CLASS].',
              'a photo of [CLASS].']
\end{lstlisting}

\begin{lstlisting}[caption=Templates for DomainNet]
templates = ['a clipart of [CLASS].',
              'a infograph of [CLASS].',
              'a painting of [CLASS].',
              'a quickdraw of [CLASS].',
              'a photo of [CLASS].',
              'a sketch of [CLASS].']
\end{lstlisting}

\noindent\textbf{More results of CAR-FT combined with WiSE-FT.} In Table~\ref{tab:imagenet} and \ref{tab:backbone}, we only display results of CAR-FT combined with WiSE-FT at optimal interpolation weight. To show more specific results, we take 10 weights uniformly from the interval [0,1] and linearly interpolate zero-shot and fine-tuned models. Such that an ID-OOD accuracy curve can be plotted to see if our CAR-FT achieves better trade-off. Figure~\ref{fig:wiseft} suggests the curve of CAR-FT combine with WiSE-FT is located above the original WiSE-FT curve. Most interpolated models of CAR-FT are better than WiSE-FT consistently. 

\bibliography{bibliography}


\end{document}